\documentclass[12pt]{article}

\usepackage{sbc-template}
\usepackage{graphicx,url}
\usepackage[utf8]{inputenc}
\usepackage[brazil]{babel}
\usepackage{booktabs}
\usepackage{tcolorbox}
\usepackage{hyperref}

\title{Usando LLMs para Programar Jogos de Tabuleiro e Variações}

\author{Álvaro Guglielmin Becker\inst{1}, Lana Bertoldo Rossato\inst{1}, Anderson Rocha Tavares\inst{1} }

\address{Instituto de Informática -- Universidade Federal do Rio Grande do Sul (UFRGS)\\
  Caixa Postal 15064 -- Porto Alegre -- RS -- Brasil
  \email{\{agbecker,lbrossato,artavares\}@inf.ufrgs.br}
}
\begin{document} 

\maketitle

\begin{abstract}
  Creating programs to represent board games can be a time-consuming task. Large Language Models (LLMs) arise as appealing tools to expedite this process, given their capacity to efficiently generate code from simple contextual information. In this work, we propose a method to test how capable three LLMs (Claude, DeepSeek and ChatGPT) are at creating code for board games, as well as new variants of existing games.

\end{abstract}
     
\begin{resumo} 
  A criação de programas que representem jogos de tabuleiro pode ser uma tarefa demorada. Large Language Models (LLMs) surgem como ferramentas interessantes para acelerar esse processo, dada sua capacidade para geração eficaz de código a partir de informações contextuais simples. Neste trabalho, propomos um método para testar a capacidade de três LLMs (Claude, DeepSeek e ChatGPT) de criar código para jogos de tabuleiro, bem como de novas variantes de jogos existentes. 
  
\end{resumo}

\section{Introdução}

Jogos de tabuleiro são um assunto importante na área de Inteligência Artificial (IA). Muitos trabalhos exploraram a criação de algoritmos para jogar estes jogos \cite{Genesereth2005ggp, strategy-game-playing}, e outros a criação de novos jogos a partir de variações sobre jogos existentes \cite{Browne2011evolutionary,lana-risk,todd2024gavel}. Em ambos os casos, é necessária a representação dos jogos estudados em código, cuja implementação pode ser uma tarefa demorada e trabalhosa. 

Nesse contexto, Large Language Models (LLMs) possuem grande potencial, dadas suas capacidades em geração de código a partir de descrições em linguagem natural \cite{Coignion2024llmcoding}. Além disso, devido à massiva quantidade de dados usada nos seus treinamentos, terão familiaridade com as regras e implementações existentes dos jogos de tabuleiro mais populares. Assim, seu uso poderia representar um ganho substancial de eficiência na implementação não só de jogos existentes, de variantes desses jogos, ou até mesmo de novos jogos descritos por analogia aos já existentes.

Neste trabalho, propomos um método para avaliar o desempenho de três LLMs (Claude 3.7 Sonnet, DeepSeekV3 e ChatGPT-4o) na tarefa de geração de código em Python para seis jogos de tabuleiro comuns, conhecidos pelos modelos. Testamos a implementação tanto dos jogos como de duas variações de cada jogo, modificando seu equipamento (tabuleiro, peças) e suas regras, para verificar a capacidade de raciocínio em alto nível dos modelos sobre aspectos dos jogos. Adicionalmente, testamos o efeito do uso da Boardwalk API \cite{boardwalk-becker}, uma plataforma própria para desenvolvimento de jogos de tabuleiro em Python que permite fácil integração com agentes de IA jogadores.

Esperamos obter uma taxa alta de sucesso, sobretudo nas implementações dos jogos sem modificações e sem uso da API, visto que os modelos estarão essencialmente reproduzindo implementações que terão certamente encontrado em seu treinamento. Havendo boa performance na implementação dos jogos com variações de regras, a descrição de jogos novos ou pouco conhecidos em termos de jogos conhecidos consolidaria-se como estratégia útil para designers de jogos e pesquisadores.


\section{Trabalhos Relacionados} \label{sec:relacionados}

A maioria dos trabalhos existentes na geração de código para jogos de tabuleiro com LLMs não utiliza linguagens de programação de propósito geral, mas sim Game Description Languages (GDLs), linguagens de descrição de jogos com gramática própria, como a Ludii GDL~\cite{ludii}. Todd et al. (2024)\nocite{todd2024gavel} fizeram fine-tuning em uma LLM para completar descrições parciais de jogos na Ludii GDL, e Tanaka e Simo-Serra (2024) \nocite{ludii-grammar-llm} usaram uma LLM para gerar descrições em Ludii a partir de prompts em linguagem natural. Entretanto, ambos os trabalhos relataram alta frequência de erros nas implementações devido à complexidade da gramática Ludii e seu relativo desconhecimento por parte dos modelos.

Em \cite{boardwalk-becker}, usamos descrições em linguagem natural de regras de jogos como prompts para geração de código em Python por LLMs, obtendo uma taxa de sucesso razoável. Além disso, introduzimos a Boardwalk API, que permite uma relativa padronização do formato do código e fácil integração com agentes de IA jogadores.

A abordagem proposta neste trabalho é nova ao se utilizar explicitamente dos conhecimentos prévios dos modelos quanto às regras dos jogos, invocando-os apenas através dos nomes dos jogos. Além disso, testamos o raciocínio em alto nível dos modelos sobre este conhecimento, conciliando o que sabem sobre as regras originais com as variações pedidas, ao invés de descrever as regras já com as modificações incorporadas.

\section{Metodologia Proposta}
Serão realizados testes com três LLMs: Claude Sonnet 3.7 \cite{claude3}, DeepSeekV3 \cite{deepseekai2025deepseekv3technicalreport} e ChatGPT-4o \cite{openai2024gpt4technicalreport}. Embora haja versões mais atualizadas destes modelos, são as mesmas versões usadas em \cite{boardwalk-becker}, permitindo assim uma comparação justa de desempenho. Os modelos serão acessados através da plataforma Poe\footnote{\url{https://poe.com}}.

Foram selecionados seis jogos de tabuleiro para teste: Jogo da Velha, Resta Um, Reversi, Moinho, Damas e Xadrez. Por serem jogos consagrados, espera-se que as LLMs tenham conhecimento suficiente sobre as regras dos jogos e implementações existentes deles em código\footnote{Como verificação primária desta hipótese, pedimos a cada modelo que descrevesse as regras de cada jogo, e todas foram fornecidas corretamente.}. Para cada jogo, serão pedidas três variações: uma original, uma mudando o equipamento (formato do tabuleiro ou distribuição de peças) e uma mudando as regras em si (condições de vitória ou funcionamento de peças).

A cada modelo, para cada jogo, serão pedidas duas formas de implementação em Python: uma com a Boardwalk API, e outra sem (implementação independente). Nos testes com a API, será fornecida apenas a documentação da API, e não seu código fonte. Da mesma forma que foi feito em \cite{boardwalk-becker}, isso permitiria verificar se o uso da API atrapalha na geração de código, embora enriqueça o resultado final devido à padronização do código e interface para jogadores humanos e agentes de IA. No total, serão executados 108 testes (3 LLMs, 6 jogos, 3 variações, 2 formas de implementação). 

\begin{figure}[ht]
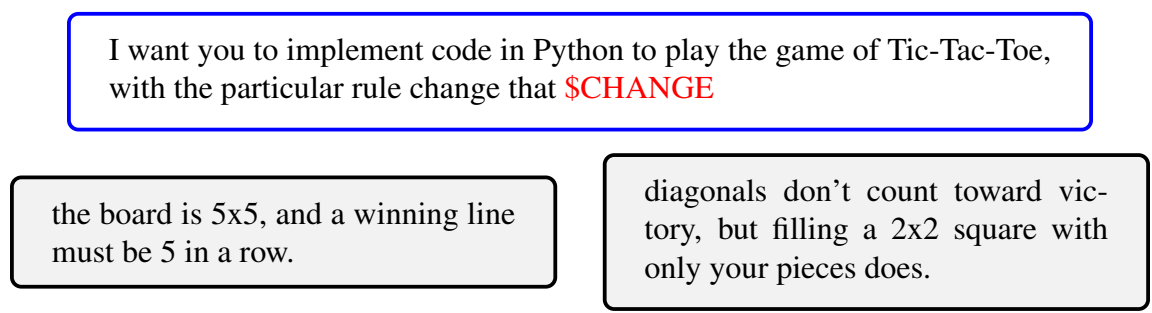

\centering
\begin{tcolorbox}[colback=white, colframe=blue, width=0.9\linewidth]
    I want you to implement code in Python to play the game of Tic-Tac-Toe, with the particular rule change that \color{red}\$CHANGE
\end{tcolorbox}

\begin{minipage}{0.48\linewidth}
    \begin{tcolorbox}[colback=gray!10, colframe=black]
        the board is 5x5, and a winning line must be 5 in a row.
    \end{tcolorbox}
\end{minipage}
\hfill
\begin{minipage}{0.48\linewidth}
    \begin{tcolorbox}[colback=gray!10, colframe=black]
        diagonals don’t count toward victory, but filling a 2x2 square with only your pieces does.
    \end{tcolorbox}
\end{minipage}
\caption{Prompt usado para a implementação independente do Jogo da Velha com variação de regras. A expressão ``\$CHANGE'' é substituída pela descrição da regra modificada nos retângulos cinza.}
\label{fig:prompts}
\end{figure}

Serão usados prompts padronizados em todos os testes. Estes diferirão dependendo do tipo de teste (independente ou com API, e jogo original ou variação); entretanto, testes na mesma categoria usarão o mesmo prompt, mudando apenas o nome do jogo e a mudança nas regras, se aplicável. A Figura \ref{fig:prompts} mostra um exemplo de prompt para variações de regra.

Os códigos gerados serão avaliados por \textit{playtest} manual, executando o programa e jogando partidas representativas do jogo para verificar a implementação. Os resultados serão quantificados pela ocorrência de erros: de sintaxe de Python, no uso da API, na movimentação das peças, nas condições de término e vitória, nos efeitos das jogadas (e.g. capturas e promoções), na formatação do tabuleiro, e na ordem de ação dos jogadores. A ocorrência de erro será tratada binariamente para cada tipo, ou seja, ou uma instância apresenta dado tipo de erro, ou não apresenta. As implementações perfeitas, sem ocorrência de nenhum tipo de erro, serão tidas como maior métrica de sucesso para um modelo. De modo similar, serão identificadas implementações que apresentem erros que impossibilitem a execução do jogo como medida de fracasso.

\section{Resultados Esperados, Conclusão e Trabalhos Futuros}
Neste trabalho, detalhamos a metodologia que usaremos para avaliar a capacidade de três LLMs de implementarem código em Python para jogos de tabuleiro, baseado no seu conhecimento prévio sobre eles. Também testaremos a capacidade dos modelos de raciocinar em alto nível sobre as regras do jogo implementado, pedindo variações dos jogos descritas em linguagem natural. Adicionalmente, verificaremos os efeitos do uso da Boardwalk API nos códigos gerados.

Como extensão de \cite{boardwalk-becker}, os resultados deste estudo se somariam aos daquele na verificação de quais abordagens de prompting garantem maior taxa de sucesso na geração de código. Resultados positivos sobretudo nos testes com variações de regras representariam maior capacidade para a geração de código para jogos obscuros ou inéditos. Esperamos que o Claude novamente demonstre o melhor desempenho dentre os três modelos. Entretanto, é possível que o sucesso médio de todos os modelos seja maior, visto que os prompts usados neste trabalho são mais simples e admitem maior de uso de conhecimento prévio dos jogos pelos modelos em relação à informação nova contida no prompt.

Trabalhos futuros podem abordar aspectos que ficariam em aberto na proposta, como a execução automatizada de playtests e a implementação de jogos completamente novos, fora do conhecimento prévio dos LLMs.

\section*{Agradecimentos}
Este estudo foi financiado em parte pela Coordenação de Aperfeiçoamento de Pessoal de Nível Superior - Brasil (CAPES) - Código de Financiamento 001 e pelo Conselho Nacional de Desenvolvimento Científico e Tecnológico (CNPq).

\bibliographystyle{sbc}
\bibliography{sbc-template}

\end{document}